\documentclass[conference]{IEEEtran}
\IEEEoverridecommandlockouts
\usepackage{cite}
\usepackage{amsmath,amssymb,amsfonts}
\usepackage{algorithmic}
\usepackage{graphicx}
\usepackage{textcomp}
\usepackage{xcolor}

\usepackage{times}
\usepackage{epsfig}
\usepackage{graphicx}
\usepackage{amsmath}
\usepackage{amssymb}
\usepackage{booktabs}
\usepackage{multirow}
\usepackage{morewrites}
\usepackage{makecell}
\usepackage{colortbl}
\usepackage{comment}
\usepackage{amsmath}
\usepackage{microtype}

\usepackage[hidelinks]{hyperref}

\usepackage{graphicx}

\usepackage{hyperref} 
\usepackage{bm}
\usepackage{amssymb} 
\usepackage{float}
\usepackage{array}
\usepackage{tabularx}
\usepackage{subcaption} 
\usepackage{tikz}
\usepackage{listings}
\usepackage[affil-it]{authblk}
\usepackage{adjustbox}
\usepackage[flushmargin]{footmisc} %
\makeatletter
\renewcommand{\footnoterule}{%
  \kern-3pt %
  \hrule width 0.5\columnwidth height 0.5pt
  \kern 2.5pt 
}
\makeatother

\def\BibTeX{{\rm B\kern-.05em{\sc i\kern-.025em b}\kern-.08em
    T\kern-.1667em\lower.7ex\hbox{E}\kern-.125emX}}
\begin{document}

\title{GraphiContact: Pose-aware Human-Scene Robust Contact Perception for Interactive Systems}

\author{
     Xiaojian Lin$^{1}$, 
     Yaomin Shen$^{2}$, 
     Junyuan Ma$^{1}$, 
     Yujie Sun$^{3}$, 
     Chengqing Bu$^{1}$, 
     Wenxin Zhang$^{4}$, \\
     {\normalfont Zongzheng Zhang$^{1}$, 
     Hao Fei$^{5}$, 
     Lei Jin$^{3}$\textsuperscript{*}, 
     Hao Zhao$^{1}$\textsuperscript{*}\thanks{* are equally corresponding authors.}} \\
     
     \emph{$^{1}$Tsinghua University, China} \\
     \emph{$^{2}$XR System Application Research Center, Nanchang Research Institute, Zhejiang University, China} \\ 
     \emph{$^{3}$Beijing University of Posts and Telecommunications, China} \\
     \emph{$^{4}$University of Chinese Academy of Sciences, China} \\
     \emph{$^{5}$National University of Singapore, Singapore} \\
     {\small \tt jinlei@bupt.edu.cn, zhaohao@air.tsinghua.edu.cn}
}

\maketitle

\begin{abstract}
Monocular vertex-level human-scene contact prediction is a fundamental capability for interactive systems such as assistive monitoring, embodied AI, and rehabilitation analysis. In this work, we study this task jointly with single-image 3D human mesh reconstruction, using reconstructed body geometry as a scaffold for contact reasoning. Existing approaches either focus on contact prediction without sufficiently exploiting explicit 3D human priors, or emphasize pose/mesh reconstruction without directly optimizing robust vertex-level contact inference under occlusion and perceptual noise. To address this gap, we propose GraphiContact, a pose-aware framework that transfers complementary human priors from two pretrained Transformer encoders and predicts per-vertex human-scene contact on the reconstructed mesh. To improve robustness in real-world scenarios, we further introduce a Single-Image Multi-Infer Uncertainty (SIMU) training strategy with token-level adaptive routing, which simulates occlusion and noisy observations during training while preserving efficient single-branch inference at test time. Experiments on five benchmark datasets show that GraphiContact achieves consistent gains on both contact prediction and 3D human reconstruction. Our code, based on the GraphiContact method, provides comprehensive 3D human reconstruction and interaction analysis, and will be publicly available at \href{https://github.com/Aveiro-Lin/GraphiContact}{https://github.com/Aveiro-Lin/GraphiContact}.
\end{abstract}


\section{Introduction}

Monocular vertex-level human-scene contact prediction is an important capability for interactive systems, because physical contact signals support safety-aware assistance, embodied AI, and rehabilitation analysis. In this work, we study this task jointly with single-image 3D human mesh reconstruction, so that reconstructed body geometry can serve as a scaffold for reasoning about where the human body touches the environment. Compared with coarse interaction recognition, vertex-level contact prediction requires sensitivity to subtle body-surface cues while remaining robust to occlusion, pose variation, and perceptual noise. These challenges are common in real-world human–computer interaction (HCI) settings, where local appearance alone is often insufficient for reliable contact inference.

\begin{figure}[t]
\begin{center}
   \includegraphics[width=0.9\columnwidth]{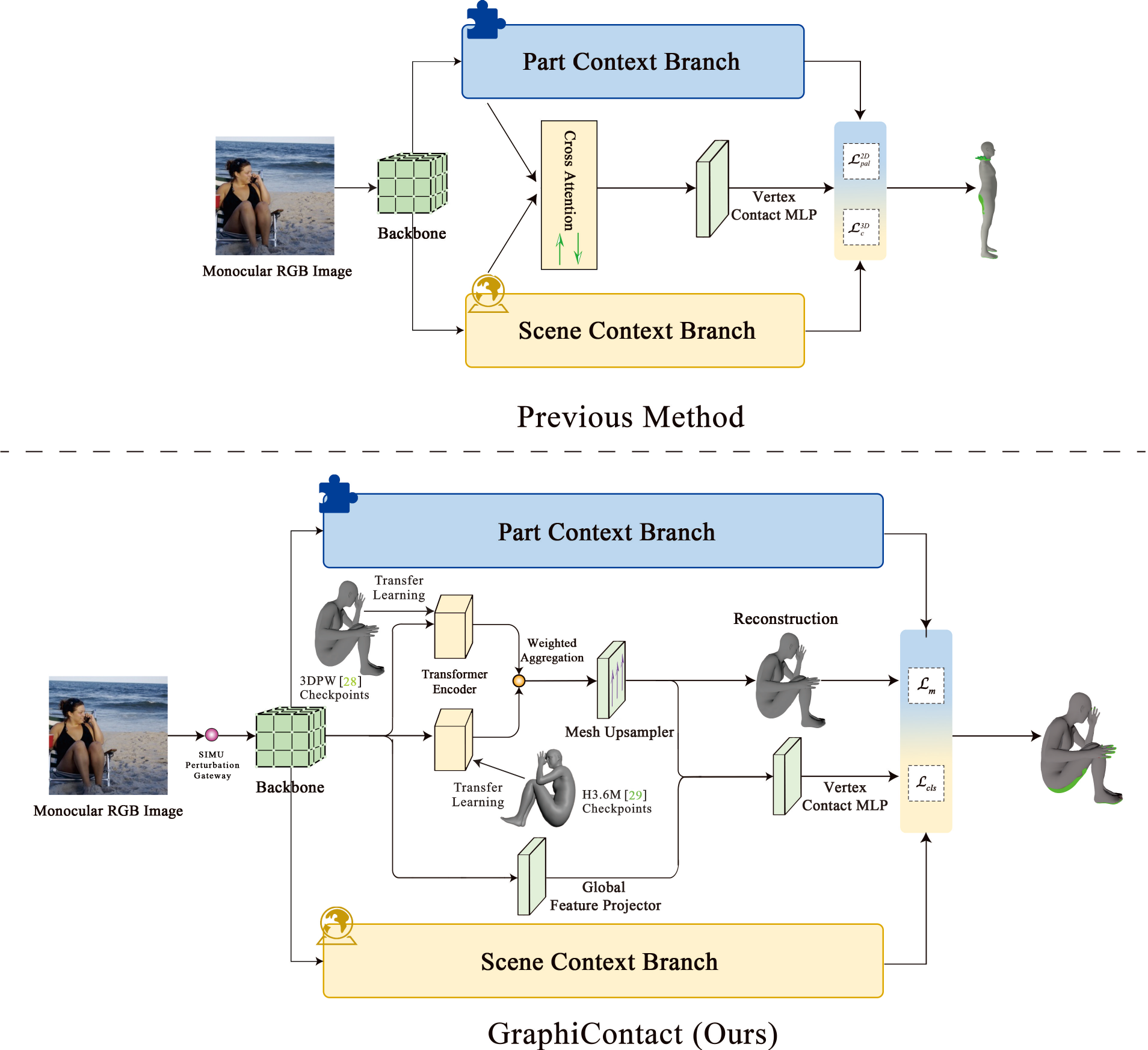}
\end{center}
\vspace{-5pt}
\caption{Motivation of GraphiContact. GraphiContact augments reconstructed human geometry with pose-aware priors and robustness training to improve contact prediction over a representative prior method~\cite{tripathi2023deco}.}
\vspace{-5pt}
\label{fig:Compare with Previous}
\end{figure}

Existing work related to our problem falls into three directions. First, contact-focused methods study specific contact types, such as hand contact~\cite{narasimhaswamy2020detecting}, foot placement~\cite{rempe2021humor}, self-contact~\cite{muller2021self}, and person-to-person interaction~\cite{fieraru2020three}. Second, broader human-scene or human-object methods extend contact reasoning to the full body~\cite{chen2023detecting,huang2022capturing}, but often lack explicit 3D human priors for robust vertex-level inference under occlusion and noise. Third, scene-aware reconstruction methods leverage GCNNs~\cite{kolotouros2019convolutional}, Transformers~\cite{vaswani2017attention}, and Graphormer-style priors~\cite{lin2021mesh} to improve pose or mesh recovery, yet do not directly optimize human-scene contact prediction.

Motivated by these gaps, we propose GraphiContact, a pose-aware framework for monocular vertex-level human-scene contact prediction jointly with single-image 3D human mesh reconstruction. As shown in Fig.~\ref{fig:Compare with Previous}, GraphiContact transfers complementary reconstruction priors from two pretrained Transformer encoders and predicts per-vertex contact on the reconstructed mesh, enabling contact reasoning and mesh recovery to reinforce each other. To improve robustness under real-world degradations, we further introduce a Single-Image Multi-Infer Uncertainty (SIMU) training strategy with token-level adaptive routing, which exposes the model to occlusion, masking, and noisy observations during training while retaining efficient single-branch inference at test time.

\begin{itemize}
    \item We present GraphiContact for monocular vertex-level human-scene contact prediction and single-image 3D mesh reconstruction.
    
    \item GraphiContact transfers complementary human priors from two pretrained Transformer encoders and uses SIMU with token-level adaptive routing for robust single-branch inference.
    
    \item Results on five benchmarks show consistent gains over prior contact-prediction and reconstruction methods.

\end{itemize}

\section{Related Work}

Our work lies at the intersection of contact prediction and single-image human reconstruction. Unlike pose/mesh-only methods, we explicitly optimize vertex-level human-scene contact on the reconstructed mesh. Unlike contact-only pipelines, we inject complementary 3D human priors from pretrained reconstruction backbones and design robustness training for occlusion and noisy observations. Therefore, GraphiContact is not a mesh-only reconstruction model with an auxiliary output; its primary goal is robust per-vertex contact inference, while mesh reconstruction serves as the geometric scaffold and auxiliary supervision.

\noindent\textbf{Human-Scene Contact Modeling.}  
Early methods, such as HOT~\cite{chen2023detecting} and~\cite{yang2021cpf}, focused on inferring 2D contact and 2.5D images~\cite{brahmbhatt2020contactpose}. Initial research primarily targeted contact detection for specific body parts, such as hands~\cite{brahmbhatt2020contactpose} and feet~\cite{rempe2021humor}, limiting their application to interactions between individual body parts and objects.  
More recent approaches have progressed to modeling full-body contacts within 3D scenes. For example, HULC~\cite{shimada2022hulc} employs contact-aware motion optimization to capture dense contact, while CHORE~\cite{xie2022chore} uses implicit surface learning to reconstruct contact from monocular images. However, these methods rely heavily on geometric constraints and face difficulties in detecting accurate contact points due to limited scene understanding.  
Works like DECO~\cite{tripathi2023deco} and BSTRO~\cite{huang2022capturing} address some of these issues but fail to fully utilize 3D information and incorporate pose relevance. In contrast, GraphiContact targets monocular vertex-level human-scene contact prediction and explicitly injects 3D human reconstruction priors to improve robustness under occlusion and noisy observations.

\noindent\textbf{Scene-Aware Human Reconstruction.} Scene context provides valuable cues for human reconstruction. Recent works leverage scene information to constrain human poses~\cite{yang2024lemon}, guide motion estimation~\cite{shimada2022hulc,liu2025frequency,wang2023himore}, or improve contact modeling~\cite{xie2022chore,wang2025factorization}. However, most existing approaches~\cite{zhu2021pseudo,zhao2020temporally,yu2025towards,xia2025reconstructing} overlook semantic scene understanding. Different from these methods, we use single-image human reconstruction as a geometric scaffold for per-vertex contact prediction rather than treating reconstruction as the sole endpoint.

\noindent\textbf{Advances in Human-Scene Interaction Understanding.} Modeling human-scene interactions has attracted increasing attention in recent years. PROX~\cite{hassan2019resolvingprox} improves pose estimation by penalizing intersections using scene constraints. MoCapDeform~\cite{li2022mocapdeform} jointly models human motion and scene deformation. LEMON~\cite{yang2024lemon} investigates semantic and geometric correlations for interaction modeling. PICO~\cite{PICO} models 3D human-object interactions, offers a promising framework for reconstructing human-object contacts in the wild. Although these methods model richer human-scene interactions, they do not directly formulate robust vertex-level contact prediction under occlusion and perceptual noise.

\section{Method}

\vspace{-4pt}
\subsection{Overview}

\noindent\textbf{Overall Framework.} As shown in Fig.~\ref{fig:overview of architecture}, GraphiContact consists of two primary branches. The \textbf{Mesh \& Pose Reconstruction Branch} (Fig.~\ref{fig:overview of architecture} (a)) utilizes our \textbf{SIMU} Strategy and transfer learning, which incorporates pose-aware features through uncertainty strategy to enhance contact point prediction and human reconstruction. This branch processes monocular RGB input via transformer encoders with grid features, global vectors, and query tokens. The \textbf{Context Integration Branch} (Fig.~\ref{fig:overview of architecture} (b)) supports the contact prediction and reconstruction tasks by integrating semantic and body-part context through specialized branches.

\label{subsection:overview}
\vspace{-2pt}

\subsection{GraphiContact Method}

\begin{figure*}

   \centering
   \includegraphics[width=0.66\linewidth]{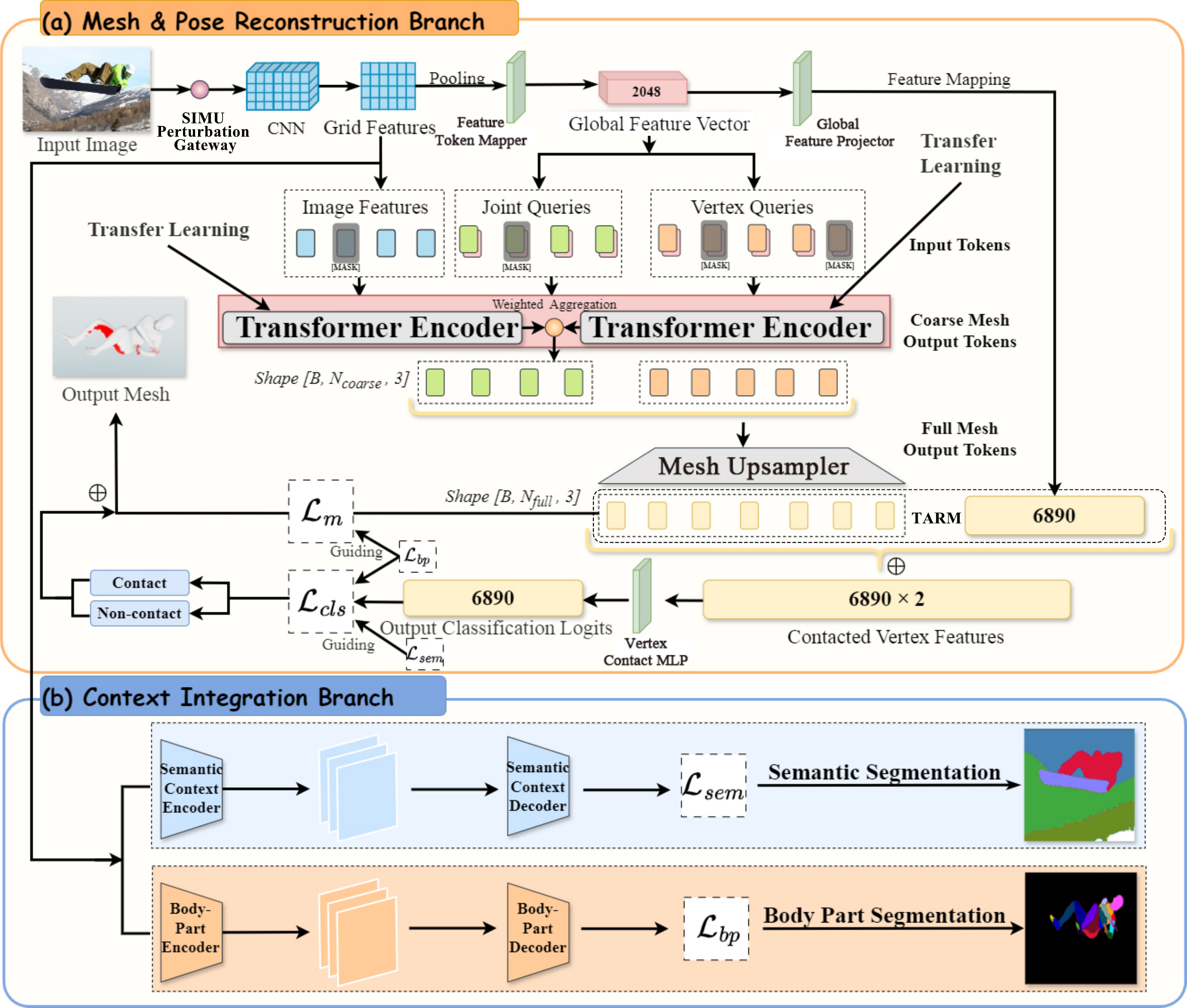}
   \vspace{-5pt}
   \caption{Comprehensive workflow of the GraphiContact method for 3D human mesh reconstruction and contact prediction with context integration. (a) The Mesh and Pose Reconstruction Branch processes an input image, which first undergoes SIMU Perturbation Gateway for SIMU Strategy, to extract features using a pre-trained CNN. These features are tokenized and passed through two Transformer Encoders, generating coarse and full mesh tokens. The mesh is refined with the mesh reconstruction loss \(\mathcal{L}_m\) and contact classification loss \(\mathcal{L}_{\text{cls}}\), determining vertex-environment contact. (b) The Context Integration Branch processes semantic and body-part context to enhance segmentation. It consists of two modules: the Semantic Context Encoder-decoder, guided by $\mathcal{L}_{\text{sem}}$, and the Body-Part Context Encoder-decoder, guided by $\mathcal{L}_{\text{bp}}$. These outputs improve contact point prediction in complex human-environment interactions.}
   \label{fig:overview of architecture}
\end{figure*}

\noindent\textbf{Mesh \& Pose Reconstruction Branch.}  
This branch focuses on reconstructing 3D human meshes while predicting contact points. As illustrated in Fig.~\ref{fig:overview of architecture}, a monocular RGB image is first processed by SIMU perturbation gateway which we will discuss in the next section, and then passed to a pre-trained CNN to extract grid features and a global feature vector. These features are tokenized into image features, joint queries, and vertex queries, which are fed into 2 Transformer Encoders. The encoders employs multi-head self-attention and graph residual blocks to capture global dependencies and local spatial relationships effectively. Using transfer learning, we incorporated the pre-trained weights of the Mesh Graphormer model~\cite{lin2021mesh} for 3D human reconstruction by introducing two transformer encoders, trained on the training sets of 3DPW~\cite{vonMarcard2018} and Human3.6M~\cite{ionescu2014human3} datasets. After passing through the two encoders, a weighted aggregation mechanism was applied to integrate prior knowledge for human reconstruction. The two Transformer encoders share the same tokenization and output aligned vertex tokens; we fuse them using a fixed scalar weighting,
$\tilde{\mathbf{m}}_{v}=1.0\,\mathbf{m}^{(3DPW)}_{v}+0.1\,\mathbf{m}^{(H3.6M)}_{v}$,
and the rationale for this fixed weighting is provided in the supplementary material.

In this paper, we use the \textbf{SMPL}~\cite{loper2015smpl} model as a parametric template for human body reconstruction. For detailed information about the template, please refer to the original paper we cited. The final outputs include a coarse mesh, refined by a mesh upsampler to produce a high-resolution mesh, along with contact probabilities for each vertex. The mesh reconstruction loss \(\mathcal{L}_m\) aligns the predicted 3D mesh with the ground truth, while the contact classification loss \(\mathcal{L}_{\text{cls}}\) identifies human-environment interaction points. For details of the loss functions, please refer to Sec.~\ref{subsection:loss}.

\noindent\textbf{Context Integration Branch.}  
This branch enhances contact predictions by incorporating both semantic and body-part level information. The semantic branch captures environmental features, such as nearby objects and ground planes, and is optimized using the semantic loss $\mathcal{L}_{\text{sem}}$. The body-part branch focuses on segmenting body parts essential for interactions and is guided by the body-part loss $\mathcal{L}_{\text{bp}}$. By combining environmental and anatomical cues with mesh outputs, these decoders refine the predicted contact regions. The two losses in the Context Integration Branch guide the two losses in the Reconstruction Branch and contribute to the optimization of the overall loss. This will be discussed in detail in the Sec.~\ref{subsection:loss}.
\label{subsection:GM}
\vspace{-0.3em}

\subsection{Single-Image Multi-Infer Uncertainty (SIMU) Strategy}  
While introducing pose-aware features through transfer learning improves human contact point prediction, challenges remain in understanding complex poses and achieving multi-scale feature fusion. These limitations become particularly evident when individuals are in complex postures or exhibit multiple contact points with the environment. 

To address these challenges, we propose the SIMU Strategy, which enhances robustness for 3D human contact point prediction and mesh reconstruction from monocular images by leveraging perturbation-induced prediction variability (i.e., ``uncertainty'') during multi-path training, rather than aiming for calibrated uncertainty estimation. As described in the Fig.~\ref{fig:overview of SIMU} and \textbf{Algorithm Table} in the supplementary materials, we propose a multi-inference path design during training, creating four distinct paths by injecting controllable perturbations at the feature level of a single input image. Techniques such as \textit{spatial dropout}, \textit{embedding noise injection}, and \textit{token-level masking} are used to mimic perceptual variations (e.g., occlusions, lighting changes, blur, low contrast), allowing each path to reflect different uncertainty aspects.

Unlike simple averaging across inference paths, we introduce a \textbf{Token-wise Adaptive Routing Module (TARM)} that assigns vertex-specific weights to each perturbed input. \textbf{TARM} assigns vertex-specific weights to each perturbed input, enabling adaptive feature fusion. Given the features $\{m_v^{(i)}\}_{i=1}^N$ at vertex $v$ from $N$ perturbations, TARM computes attention scores for each perturbation:

\begin{align}   
\alpha_v^{(i)} = \text{softmax} \left( \mathbf{w}^\top \cdot \varphi(m_v^{(i)}) \right), \quad i = 1, \dots, N.
\end{align}

The $\mathbf{w}^\top$ is the transpose of a learnable weight vector and $\varphi(m_v^{(i)})$ represents the transformation applied to the feature $m_v^{(i)}$ at vertex $v$. The features are then fused as:

\begin{align}
\hat{m}_v = \sum_{i=1}^N \alpha_v^{(i)} m_v^{(i)}.
\end{align}

This fusion mechanism allows the model to focus on informative paths while down-weighting less reliable ones. For example, under occlusion, dropout paths are given lower weight, while higher visibility paths are prioritized. The adaptive fusion supports joint modeling of epistemic and aleatoric uncertainties. During training, the fused features are used to supervise both contact prediction and mesh reconstruction, improving robustness in challenging conditions like low visibility or occlusion.

\label{subsection:SIMU}

\subsection{Loss Function}  

During reconstruction, the loss function \(\mathcal{L}_m\) is used to minimize mesh reconstruction error by optimizing the predicted 3D coordinates of the mesh vertices. For human-environment interaction, \(\mathcal{L}_{\text{cls}}\), defined as a binary cross-entropy loss, classifies contact points. This loss determines whether each of the 6,890 vertices is in contact with the surrounding environment using a binary 0/1 classification scheme. Relying solely on reconstruction features makes it difficult to accurately capture the human and environmental context. To address this, we employ $\mathcal{L}_{\text{bp}}$ and $\mathcal{L}_{\text{sem}}$, which are segmentation metrics defined between predicted and ground-truth masks. The $\mathcal{L}_{\text{bp}}$ loss supervises the human part context, while the $\mathcal{L}_{\text{sem}}$ loss supervises the scene context. Specifically, the body-part supervision uses SMPL-style body-part masks, obtained by rendering the posed human mesh into a 2D part segmentation (or by rendering from a pseudo mesh when ground-truth meshes are unavailable). The semantic-context supervision uses semantic scene masks obtained from an off-the-shelf image segmentation model as pseudo ground truth. These losses guide the model to focus on fine-grained human details and human-environment interaction context, enhancing the overall performance of contact point prediction.

We supervise mesh reconstruction with $\mathcal{L}_m$ as an MSE regression loss on 3D vertex coordinates, use BCE for the vertex-wise contact classification loss $\mathcal{L}_{\text{cls}}$, and apply auxiliary scene/body-part segmentation supervision via $\mathcal{L}_{\text{sem}}$ and $\mathcal{L}_{\text{bp}}$ with task-appropriate multi-class segmentation losses. The exact loss formulations and weighting scheme are provided in the supplementary material (Algorithm Table).
The overall training objective for the model is defined as:
\begin{align}
    \mathcal{L}_{all} = w_m\mathcal{L}_m + w_{cls}\mathcal{L}_{\text{cls}} + w_{\text{sem}}\mathcal{L}_{\text{sem}} + w_{\text{bp}}\mathcal{L}_{\text{bp}},
\end{align}
where \(w_m = 1\), \(w_{cls} = \{1, 0.1\}\) (corresponding to the two encoders), \(w_{\text{sem}} = 1\), and \(w_{\text{bp}} = 1\).

\label{subsection:loss}

\begin{figure}[t]
\vspace{-2pt}
\begin{center}
   \includegraphics[width=0.90\columnwidth]{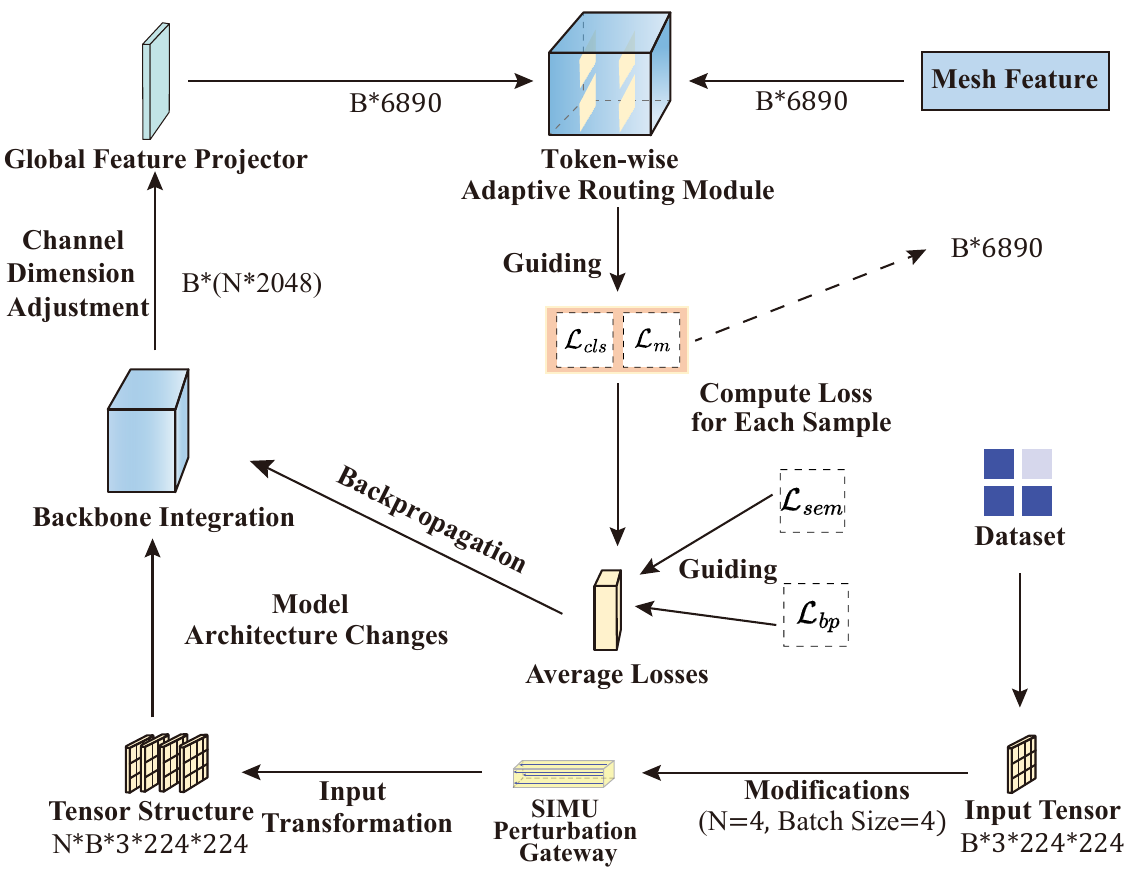}
\end{center}

\caption{Overview of the SIMU Strategy for parallelized perturbation learning and contact prediction. It transforms the input tensor, applies multi-path perturbations via the SIMU gateway, and extracts features with the backbone. TARM performs token-wise weighted fusion across paths under multi-task supervision ($\mathcal{L}_{\text{cls}}$, $\mathcal{L}_m$, $\mathcal{L}_{\text{sem}}$, $\mathcal{L}_{\text{bp}}$), and the aggregated loss is backpropagated to improve robustness and context-aware contact estimation.}

\label{fig:overview of SIMU}
\end{figure}

\begin{table*}[!ht]

 \centering
 \caption{
     Comparison of GraphiContact with SOTA Models on RICH~\cite{huang2022capturing}, DAMON~\cite{tripathi2023deco}, and BEHAVE~\cite{bhatnagar2022behave}. 
}
\vspace{-0.5em}
\label{table:compare-sota}
\setlength{\tabcolsep}{0.6mm}
\begin{tabular}{c|cccc|cccc|cccc}   
\Xhline{2\arrayrulewidth}
\multirow{2}{*}{\bf Methods} & \multicolumn{4}{c|}{\bf RICH~\cite{huang2022capturing}} & \multicolumn{4}{c|}{\bf DAMON~\cite{tripathi2023deco}} & \multicolumn{4}{c}{\bf BEHAVE~\cite{bhatnagar2022behave}} \\ 
\cline{2-13}
&\hspace{0.2em} \bf Precision $\kern-0.4em\uparrow$ & \hspace{0.2em} \bf Recall $\kern-0.4em\uparrow$ & \hspace{0.2em} \bf F1 $\kern-0.4em\uparrow$ & \hspace{0.2em} \bf geo. (cm) $\kern-0.4em\downarrow$ 
&\hspace{0.2em} \bf Precision $\kern-0.4em\uparrow$ & \hspace{0.2em} \bf Recall $\kern-0.4em\uparrow$ & \hspace{0.2em} \bf F1 $\kern-0.4em\uparrow$ & \hspace{0.2em} \bf geo. (cm) $\kern-0.4em\downarrow$ 
&\hspace{0.2em} \bf Precision $\kern-0.4em\uparrow$ & \hspace{0.2em} \bf Recall $\kern-0.4em\uparrow$ & \hspace{0.2em} \bf F1 $\kern-0.4em\uparrow$ & \hspace{0.2em} \bf geo. (cm) $\kern-0.4em\downarrow$ \\ \hline
\rowcolor{gray!13} BSTRO~\cite{huang2022capturing} & 0.65 & 0.66 & 0.63 & 18.39 & 0.51 & 0.53 & 0.46 & 38.06 & 0.13 & 0.03 & 0.04 & 50.45 \\ 
POSA$^{\text{PIXIE}}$~\cite{hassan2021populating, feng2021collaborative} & 0.31 & 0.69 & 0.39 & 21.16 & 0.42 & 0.34 & 0.31 & 33.00 & 0.11 & 0.07 & 0.06 & 54.29 \\ 
\rowcolor{gray!13} POSA$^{\text{GT}}$~\cite{hassan2021populating, feng2021collaborative} & 0.37 & 0.76 & 0.46 & 19.96 & - & - & - & - & 0.10 & 0.09 & 0.06 & 55.43 \\ 
DECO~\cite{tripathi2023deco} & 0.71 & 0.76 & 0.70 & 17.92 & 0.64 & 0.57 & 0.55 & 21.32 & 0.25 & \underline{0.21} & 0.18 & 46.33 \\ 
\rowcolor{gray!13} DECO$+\mathcal{L}_{pal}^{2D}$~\cite{tripathi2023deco} & 0.71 & 0.74 & 0.69 & 10.46 & 0.65 & 0.57 & 0.55 & 21.88 & 0.27 & 0.18 & 0.18 & 44.51 \\ 
LEMON~\cite{yang2024lemon} & 0.72 & 0.78 & 0.75 & 9.27 & 0.64 & \underline{0.58} & 0.55 & 20.91 & 0.28 & \underline{0.21} & 0.19 & 39.29 \\ 
\rowcolor{gray!13} CONTHO~\cite{nam2024joint} & \underline{0.75} & \underline{0.81} & \underline{0.78} & \underline{6.43} & \underline{0.66} & 0.57 & \underline{0.58} & \underline{19.01} & \underline{0.31} & \underline{0.21} & \underline{0.23} & \underline{23.75} \\ 
PICO~\cite{PICO} & \underline{0.75} & 0.80 & 0.77 & 6.88 & 0.64 & \underline{0.58} & 0.57 & 20.16 & 0.30 & 0.20 & 0.21 & 25.64 \\ 

\hline
\rowcolor{gray!13} GraphiContact (Ours)  & \bf 0.76 & \bf 0.95 & \bf 0.85 & \bf 3.22 & \bf 0.70 & \bf 0.63 & \bf 0.66 & \bf 15.69 & \bf 0.36 & \bf 0.34 & \bf 0.35 & \bf 12.68 \\ \Xhline{2\arrayrulewidth}
\end{tabular}

\vspace{2mm}
  \textbf{Bold} indicates the best performance for each metric, while \underline{underline} denotes the second-best result.
\vspace{-1em}  
\end{table*}

\begin{table*}[!ht]
\vspace{1em}
\centering
\caption{
     Ablation Study of GraphiContact on RICH~\cite{huang2022capturing}, DAMON~\cite{tripathi2023deco}, and BEHAVE~\cite{bhatnagar2022behave}. 
}
\vspace{-0.5em}
\label{table:ablation N values}
\setlength{\tabcolsep}{0.9mm}
\resizebox{\textwidth}{!}{  
\begin{tabular}{c|cccccc|cccc|cccc|cccc}   
\Xhline{2\arrayrulewidth}
\multirow{2}{*}{\bf Method} & \multirow{2}{*}{\bf H3.6M} & \multirow{2}{*}{\bf 3DPW} & \multirow{2}{*}{\bf TARM} & \multirow{2}{*}{\bf Dropout} & \multirow{2}{*}{\bf Noise} & \multirow{2}{*}{\bf Masking} & \multicolumn{4}{c|}{\bf RICH~\cite{huang2022capturing}} & \multicolumn{4}{c|}{\bf DAMON~\cite{tripathi2023deco}} & \multicolumn{4}{c}{\bf BEHAVE~\cite{bhatnagar2022behave}} \\ 
\cline{8-19}
& & & & & & & \bf Precision $\kern-0.4em\uparrow$ & \bf Recall $\kern-0.4em\uparrow$ & \bf F1 $\kern-0.4em\uparrow$ & \bf geo. (cm) $\kern-0.4em\downarrow$ 
& \bf Precision $\kern-0.4em\uparrow$ & \bf Recall $\kern-0.4em\uparrow$ & \bf F1 $\kern-0.4em\uparrow$ & \bf geo. (cm) $\kern-0.4em\downarrow$ 
& \bf Precision $\kern-0.4em\uparrow$ & \bf Recall $\kern-0.4em\uparrow$ & \bf F1 $\kern-0.4em\uparrow$ & \bf geo. (cm) $\kern-0.4em\downarrow$ \\ \hline
\rowcolor{gray!13} (1)&   &  &  &  &  &  &  0.73 & 0.75 & 0.74 & 8.94 &  0.64 & 0.59 & 0.57 & 21.98 & 0.32 & 0.22 & 0.25 & 14.58 \\ 
(2) & \checkmark &  &  &  &  &  & 0.74 & 0.76 & 0.75 & 8.11 & 0.64 & 0.60 & 0.57 &  21.72 & 0.32 & 0.22 & 0.25 & 14.19 \\ 
\rowcolor{gray!13} (3)& \checkmark  & \checkmark &  &  &  &  & \bf 0.76 & 0.76 & 0.76 & 7.19 &  0.66 & 0.60 & 0.58 & 20.12 & 0.34 & 0.23 & 0.26 & 13.78 \\ 
(4) & \checkmark & \checkmark & \checkmark & \checkmark &  &  & \bf 0.76 & 0.90 & 0.83 & 5.80 & 0.66 & 0.59 & 0.60 &  17.06 & 0.29 & 0.29 & 0.29 & 14.42 \\ 
\rowcolor{gray!13} (5) & \checkmark & \checkmark & \checkmark & \checkmark & \checkmark &  & \bf 0.76 & 0.94 & \bf 0.85 & 3.66 &  0.69 & \bf 0.63 &  0.65 & 15.91 & 0.34 & 0.33 & 0.33 & 13.09 \\ 
GraphiContact (Ours) & \checkmark & \checkmark & \checkmark & \checkmark & \checkmark & \checkmark & \bf 0.76 & \bf 0.95 & \bf 0.85 & \bf 3.22 & \bf 0.70 & \bf 0.63 & \bf 0.66 & \bf 15.69 & \bf 0.36 & \bf 0.34 & \bf 0.35 & \bf 12.68 \\ \Xhline{2\arrayrulewidth}
\end{tabular}
}
\vspace{-1em}  
\end{table*}

\setlength{\floatsep}{0pt}
\setlength{\textfloatsep}{0pt}
\vspace{-10pt}

\begin{table}[t]
\centering
\scriptsize 
\caption{Comparison of GraphiContact with SOTA methods on 3DPW~\cite{vonMarcard2018} and Human3.6M~\cite{ionescu2014human3}
(trained without dual-encoder transfer initialization for fairness).
}
\vspace{-0.5em}
\label{table:compare-h36m-3dpw}
\resizebox{\columnwidth}{!}{%
\begin{tabular}{c|ccc|c@{\hskip 2pt}cc} 
\Xhline{1.2pt}
\multirow{2}{*}{\textbf{Methods}} & \multicolumn{3}{c|}{\bf 3DPW~\cite{vonMarcard2018}} & \multicolumn{3}{c}{\bf Human3.6M~\cite{ionescu2014human3}} \\ 
\cline{2-7}
&\bf MPVE $\downarrow$ &\bf MPJPE $\downarrow$ &\bf PA-MPJPE $\downarrow$ & &\bf MPJPE $\downarrow$ &\bf PA-MPJPE $\downarrow$ \\
\hline

\rowcolor{gray!13}GraphCMR~\cite{kolotouros2019convolutional} & $-$ & $-$ & $70.2$ && $-$ & $50.1$\\

Mesh Graphormer~\cite{lin2021mesh} & $87.7$ & $74.7$ & $45.6$ && $51.2$ & $34.5$\\
\rowcolor{gray!13}POTTER~\cite{zheng2023potter} & $87.4$ & $75.0$ & $44.8$ && $56.5$ & $35.1$\\ 
Zolly~\cite{wang2023zolly} & $76.3$ & $65.0$ & $39.8$ && $49.4$ & $32.3$\\ 
\rowcolor{gray!13}ScoreHypo~\cite{xu2024scorehypo} & \underline{$71.9$} & \underline{$61.8$} & \underline{$36.1$} && \underline{$37.4$} & $35.3$\\ 
Hulk~\cite{wang2025hulk} & $77.4$ & $66.3$ & $38.5$ && $40.3$ & \underline{$28.8$}\\ 

\rowcolor{gray!13} HSMR~\cite{xia2025reconstructing} & $-$ & $81.1$ & $51.1$ && $52.0$ & $32.1$ \\
\hline
 GraphiContact (Ours) & $\textbf{69.3}$ & $\textbf{59.1}$ & $\textbf{34.0}$ && $\textbf{35.5}$ & $\textbf{23.9}$ \\
\Xhline{1.2pt}
\end{tabular}
}

\vspace{2mm}
\textbf{Bold} indicates the best performance for each metric, while \underline{underline} denotes the second-best result.
\end{table}

\begin{table}[t]
\centering
\scriptsize 
 \caption{GraphiContact with Different Losses on the DAMON~\cite{tripathi2023deco}.}
\vspace{-0.5em}
\label{table:damon_losses}
\resizebox{\columnwidth}{!}{%
\begin{tabular}{c|cc|cccc}
\Xhline{2\arrayrulewidth}
\textbf{Model} & \textbf{$\mathcal{L}_{\text{sem}}$} & \textbf{$\mathcal{L}_{\text{bp}}$} & \textbf{\scriptsize Precision(\%) $\kern-0.4em\uparrow$} & \textbf{\scriptsize Recall(\%) $\kern-0.4em\uparrow$} & \textbf{\scriptsize F1-Score(\%) $\kern-0.4em\uparrow$} & \textbf{\scriptsize geo.(cm) $\kern-0.4em\downarrow$} \\ \hline
\rowcolor{gray!13} (1)     & & & 70.8\% & 57.1\% & 62.7\% & 15.33 \\ 
(2)       & \checkmark & & 71.2\% & 59.5\% & 63.2\% & 14.57 \\ 
\rowcolor{gray!13} (3)       & & \checkmark & \textbf{74.0\%} & 53.2\% & 61.4\% & \textbf{9.00} \\ 
GraphiContact (Ours) & \checkmark & \checkmark & 70.4\% & \textbf{63.3\%} & \textbf{65.2\%} & 15.69 \\ \Xhline{2\arrayrulewidth}
\end{tabular}
}
\vspace{0.8em}
\end{table}

\vspace{1.2em}
\section{Experiments}  
\subsection{Experiment Settings}

\noindent\textbf{Dataset.}  
For the contact prediction task, our method is evaluated on the BEHAVE~\cite{bhatnagar2022behave}, DAMON~\cite{tripathi2023deco} and RICH~\cite{huang2022capturing} datasets, which provide comprehensive annotations of human-object interactions in diverse and domestic environments, respectively. For the 3D reconstruction task, we also use the 3DPW~\cite{vonMarcard2018} dataset for in-the-wild evaluation and the Human3.6M~\cite{ionescu2014human3} dataset for controlled indoor benchmarking. The detailed train/test splits and implementation specifics of our experimental protocol are provided in the supplementary material (Section ``Dataset Splits and Evaluation Protocols'').

\noindent\textbf{Implementation Details.} For HCI tasks, inference time is particularly critical. At inference time, we perform \emph{single-path} inference ($N{=}1$), i.e., SIMU multi-path perturbations are used only during training; the inference time reported in Tab.~\ref{table:Inference Time} (see supplementary material) is measured under this single-path setting, together with the corresponding backbone parameter count. We adopt the single-path setting to keep latency low for interactive HCI scenarios, since enabling SIMU at test time would multiply the backbone forward passes and incur unnecessary overhead. Empirically, the robustness benefits of SIMU are already distilled into the model parameters during training, and single-path inference achieves a favorable accuracy--latency trade-off without requiring multi-path ensembling. Additional implementation details and evaluation protocols of \textit{GraphiContact} are provided in the supplementary material under the section \emph{Implementation Details}.


\subsection{Main Results}  

\begin{figure}

\begin{center}
   \includegraphics[width=0.45\textwidth]{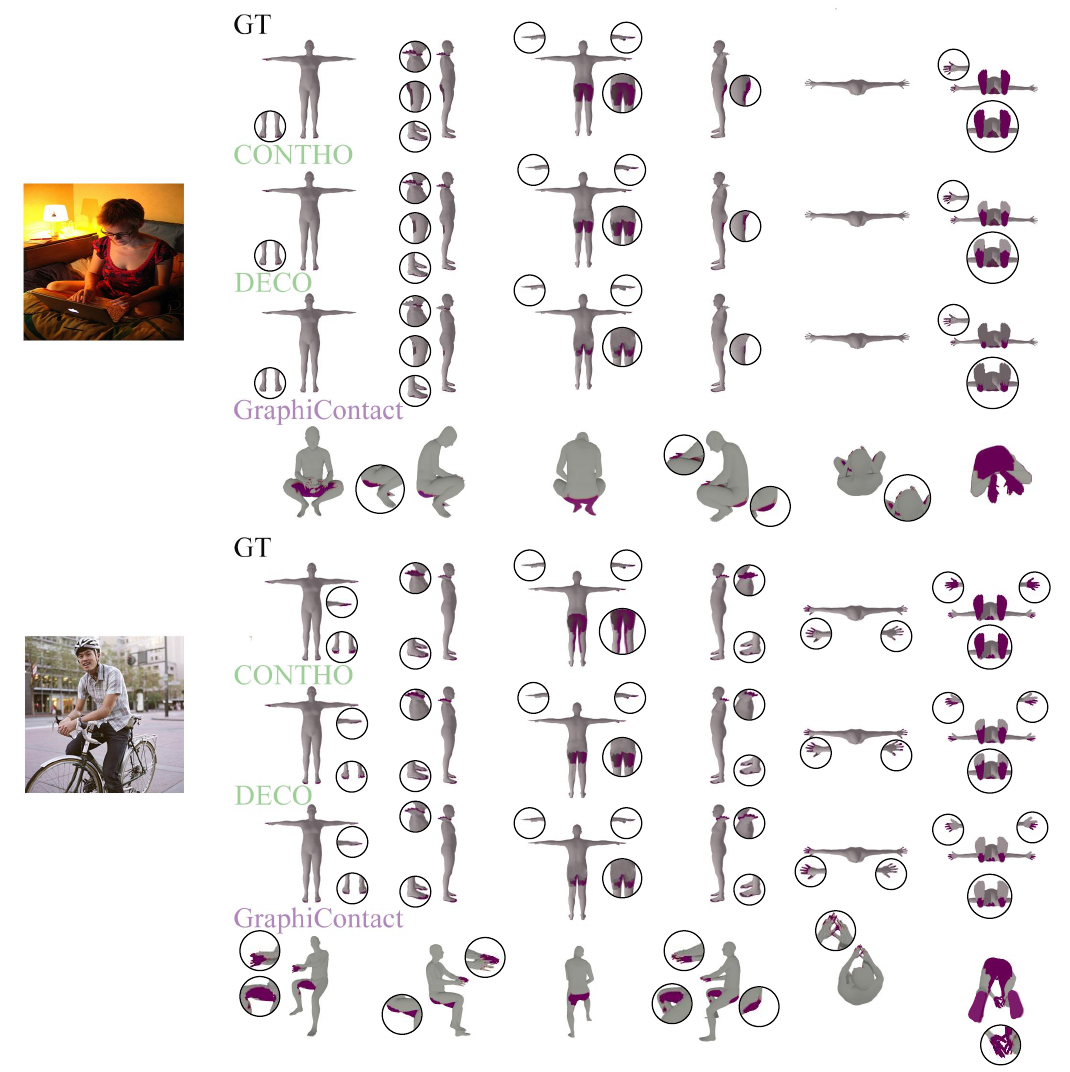}
\end{center}
\caption{Comparison of 3D human mesh reconstruction and contact prediction performance across DECO~\cite{tripathi2023deco}, CONTHO~\cite{nam2024joint}, and GraphiContact.}
\vspace{1mm}
\label{fig:results visualization}
\end{figure}

\noindent\textbf{Quantitative Comparison.}  
We conducted quantitative evaluations on the RICH~\cite{huang2022capturing}, DAMON~\cite{tripathi2023deco}, and BEHAVE~\cite{bhatnagar2022behave} datasets. Tab.~\ref{table:compare-sota} presents the quantitative results, along with comparisons to other methods. The evaluation was conducted using count-based metrics and geodesic error, following the protocol in~\cite{huang2022capturing}. Our approach achieves the best performance across all metrics, demonstrating its effectiveness and superiority. GraphiContact improves recall significantly by detecting more potential contact points, especially under challenging conditions. This is due to its pose-aware and uncertainty strategy, which enhance sensitivity. However, precision sees only marginal improvement because the model predicts more contact points, leading to some false positives. It is worth noting that the three benchmark datasets are all annotated at the vertex-level, i.e., contact labels are provided per mesh vertex rather than coarse regions, and inherently cover a wide range of challenging human-scene interactions with complex poses. Therefore, the four quantitative metrics we report in Tab.~\ref{table:compare-sota} are already the most direct instantiation of fine-grained contact prediction under such conditions. In addition, as shown in Tab.~\ref{table:compare-h36m-3dpw}, our method achieves SOTA performance on the 3DPW~\cite{vonMarcard2018} and Human3.6M~\cite{ionescu2014human3} datasets for 3D mesh reconstruction. Although none of the existing datasets provide explicit pose-complexity labels, our superior performance on two datasets well-known for their wide range of challenging poses serves as strong evidence of GraphiContact’s robustness under complex pose conditions. For fairness, the reconstruction experiments in Tab.~\ref{table:compare-h36m-3dpw} do not use the dual-prior initialization adopted in our main transfer-learning pipeline; instead, we train and evaluate the reconstruction branch without loading the 3DPW/Human3.6M pre-trained encoder checkpoints, ensuring a clean evaluation setup.

\noindent\textbf{Qualitative Comparison.}  
Fig.~\ref{fig:results visualization} presents a qualitative comparison with the DECO~\cite{tripathi2023deco}, and CONTHO~\cite{nam2024joint} method. The results show that our model achieves more accurate human contact prediction under challenging image conditions, such as occlusion by the quilt and laptop or illumination changes caused by the desk lamp in the first photo, as well as occlusion by the bicycle in the second photo.

\subsection{Ablation Study}
\vspace{0.3em}

Tab.~\ref{table:ablation N values} presents the results of our ablation study under different SIMU perturbation strategies and pre-trained weights. It is worth noting that the TARM is defined as a vertex-wise weighted fusion across $N$ inference paths, with the weights derived from attention. Therefore, when SIMU Strategy is not applied ($N=1$), the use of TARM is functionally equivalent to not using it at all. As shown, when we employ 2 weights together with four streams simultaneously—namely the original input, Dropout, Noise, and Masking, the model achieves the best performance across all three datasets. The performance improvement in GraphiContact primarily comes from SIMU Strategy rather than pose priors. SIMU enhances robustness by simulating perceptual variations like occlusion and lighting changes, which helps the model handle diverse real-world conditions. Tab.~\ref{table:damon_losses} presents the effects of different loss functions in the context integration branch on the DAMON~\cite{tripathi2023deco} dataset. The results show that $\mathcal{L}_{\text{sem}}$ improves recall, while $\mathcal{L}_{\text{bp}}$ significantly enhances precision and reduces geodesic error. This finding supports our discussion in Sec.~\ref{subsection:loss}, where $\mathcal{L}_{\text{bp}}$ is shown to encourage the model to focus on body details, leading to improved precision and lower geodesic error. In contrast, $\mathcal{L}_{\text{sem}}$ emphasizes human-environment interactions, contributing to higher recall.

\section{Conclusion}  
In this paper, we presented GraphiContact, an innovative method for 3D human reconstruction and contact point prediction. By integrating pose-aware transfer learning and SIMU Strategy, GraphiContact robustly improves contact point prediction and 3D mesh reconstruction from monocular images. Extensive experiments on multiple benchmark datasets demonstrate that GraphiContact outperforms existing SOTA methods in both human-environment contact prediction and 3D human mesh reconstruction tasks. Beyond technical accuracy, GraphiContact provides a scalable perceptual module that can support real-world HCI scenarios—such as assistive monitoring, rehabilitation tracking, and contact-aware embodied systems—where understanding physical interactions is critical. We believe this work opens new opportunities for integrating contact-aware perception into interactive AI systems.



\bibliographystyle{IEEEbib}
\bibliography{icme2026references}
\clearpage  

\twocolumn[{
\begin{center}
    {\LARGE\bf Supplementary Materials: GraphiContact: Pose-aware Human-Scene Robust Contact Perception for Interactive Systems}
    \vspace{2em}
\end{center}
}]
\section{Details of Single-Image Multi-Infer Uncertainty (SIMU) Strategy}

The \textbf{SIMU Strategy} framework enhances the robustness and uncertainty-awareness of 3D human contact point prediction and reconstruction from monocular images. The method addresses challenges arising from complex human poses and multiple contact points with the environment, particularly under challenging conditions such as occlusion and partial visibility.

The \textbf{SIMU Strategy} pipeline can be generally divided into three stages.

\noindent\textbf{1. Multi-Infer Perturbation:} After passing through the SIMU Perturbation Gateway, the original tensor with shape $(B \times 3 \times 224 \times 224)$ is restructured into a multi-inference tensor with shape $(N \times B \times 3 \times 224 \times 224)$. During training, we perform multi-infer uncertainty strategy on a single image. The first inference stream processes the original image, the second stream applies \textit{spatial dropout}, the third stream executes \textit{embedding noise injection}, and the final stream utilizes \textit{token-level masking}. The number of active inference, denoted as $N$, can be dynamically adjusted to control how many of these branches are executed. This transformation enables parallel processing of features across multi-infer streams to mimic perceptual variations, which is essential for capturing diverse aspects of the input image.

\noindent\textbf{Feature Extraction \& Adjustment:} In this stage, multi-inference streams undergo feature extraction and dimensional adjustments. Each perturbed input stream $I^{(i)}$ is passed through a shared backbone network $F_{\theta}$ to extract features $f^{(i)} \in \mathbb{R}^{B \times D}$. The extracted features from all $N$ inference paths are then concatenated along the feature dimension to form a unified representation $f \in \mathbb{R}^{B \times (N \cdot D)}$. A channel adjustment step follows, ensuring consistency in the dimensionality of the processed features. This adjustment prepares the features for subsequent global feature projection and token-wise adaptive routing, facilitating robust fusion across the multi-inference paths.

\noindent \textbf{Loss Aggregation:}  For model optimization, a weighted loss aggregation strategy is implemented, which combines different losses from each branch and minimizes parameter overhead during backpropagation. Prior to the final loss computation, as mentioned in the main text, the loss term $\mathcal{L}_{\text{sem}}$ serves as a supervisory signal for $\mathcal{L}_{\text{cls}}$, while $\mathcal{L}_{\text{bp}}$ provides guidance for both $\mathcal{L}_{\text{cls}}$ and $\mathcal{L}_m$. Although the weights of each loss term are constants rather than variables, the guiding roles between different losses still make the integration strategy worth mentioning. After these guiding effects are incorporated, the four loss terms are integrated with appropriate weighting for the final loss calculation. The detailed procedure can be found in Sec.~\ref{subsection:loss}.

A comprehensive overview of the SIMU Strategy process is provided in the following algorithm table.

\renewcommand{\arraystretch}{1.3} 
\setlength{\tabcolsep}{0pt}

\begin{table}[ht]
\centering
\setlength{\arrayrulewidth}{0.4mm} 
\begin{tabular}{@{}p{\linewidth}<\raggedright@{}} 
\hline
{\vskip 0.2em} 
  {\normalsize \textbf{Algorithm:} SIMU Strategy} \\ 
{\vskip 0.2em} \\
\hline
{\vskip 0.2em} 
\textbf{Result:} Contact probability map $\mathcal{P}$, Reconstructed mesh $\mathcal{M}$. \\[0.3em]  
 \textbf{Init:} RGB image tensor $\mathcal{I} \in \mathbb{R}^{B \times 3 \times 224 \times 224}$. \\[0.3em]
 \textbf{Step 1:} \textbf{Multi-Infer Perturbation:} Generate $N$ versions of $\mathcal{I}$:
\\
\quad $\bullet$ $\mathcal{I}^{(1)}$ = original input\\
\quad $\bullet$ $\mathcal{I}^{(2)}$ = spatial dropout\\
\quad $\bullet$ $\mathcal{I}^{(3)}$ = embedding noise injection\\
\quad $\bullet$ $\mathcal{I}^{(4)}$ = token-level masking
\\[0.3em]
 \textbf{Step 2:} \textbf{Feature Extraction:} Each $\mathcal{I}^{(i)}$ is passed through a shared backbone $\mathcal{F}_\theta$ to extract:
$$ \mathbf{f}^{(i)} = \mathcal{F}_\theta(\mathcal{I}^{(i)}), \quad \mathbf{f}^{(i)} \in \mathbb{R}^{B \times D}. $$ \\[0.3em]
 \textbf{Step 3:} \textbf{Channel Adjustment:}
$$ \mathbf{f} = \text{Concat}([\mathbf{f}^{(1)}, \ldots, \mathbf{f}^{(N)}]) \in \mathbb{R}^{B \times (N \cdot D)}. $$ \\[0.3em]
 \textbf{Step 4:} \textbf{Global Feature Projection:} Map $\mathbf{f}$ to mesh feature:
$$ \mathbf{m} \in \mathbb{R}^{B \times 6890}. $$ \\[0.3em]
 \textbf{Step 5:} \textbf{Token-wise Adaptive Routing:} 
For each vertex $v$, compute the weighted fusion across all inference paths:
\[
\begin{array}{l}
\text{(a) Attention score:} \alpha_v^{(i)} = \text{Softmax}\left(w^\top \cdot \phi(\mathbf{m}_v^{(i)})\right), \quad \text{for } i = 1, \dots, N. \\
\text{(b) Feature fusion:}\hat{\mathbf{m}}_v = \sum_{i=1}^{N} \alpha_v^{(i)} \cdot \mathbf{m}_v^{(i)}.
\end{array}
\] \\[0.3em]
 \textbf{Step 6:} \textbf{Losses:} Define the following loss terms:
\[
\begin{array}{l}
\text{Mesh reconstruction loss:}  \mathcal{L}_m = \text{MSE}(\hat{\mathcal{M}}, \mathcal{M}^{GT}). \\
\text{Contact classification loss:}  \mathcal{L}_{cls} = \text{BCE}(\hat{\mathcal{P}}, \mathcal{P}^{GT}). \\
\text{Auxiliary losses:}  \mathcal{L}_{sem}, \ \mathcal{L}_{bp}  \text{(semantic and body-part supervision).}
\end{array}
\] \\[0.3em]
 \textbf{Step 7:} \textbf{Loss Aggregation:} Normalize the auxiliary losses and compute the total loss:
\[
\begin{array}{l}
\text{Final loss:} \\
\quad \mathcal{L}_{all} = w_m\mathcal{L}_m + w_{cls}\mathcal{L}_{\text{cls}} + w_{\text{sem}}\mathcal{L}_{\text{sem}} + w_{\text{bp}}\mathcal{L}_{\text{bp}}.
\end{array} 
\]
The \(w_m = 1\), \(w_{cls} = \{1, 0.1\}\) (corresponding to the two encoders), \(w_{\text{sem}} = 1\), and \(w_{\text{bp}} = 1\).
\\[0.3em]
 \textbf{Step 8:} \textbf{Backpropagation:} Gradients from $\mathcal{L}_{all}$ update all learnable parameters in the shared backbone, global projection, adaptive routing, and task-specific heads. \\[0.3em]

\hline
\end{tabular}
\end{table}

\section{Model Architecture}
\noindent\textbf{Graphormer Encoder.}  
In the main text, we use the term \textbf{Transformer Encoder} to describe the fundamental architecture of our model. This choice reflects our effort to simplify and unify the terminology by referring to the most general form of the encoder architecture. However, the actual implementation of our GraphiContact Model builds upon the \textbf{Graphormer Encoder}. This encoder is a specialized variant of the Transformer Encoder designed specifically for graph-based tasks. To provide a comprehensive understanding of our approach, we describe the Graphormer Encoder in this supplementary material, highlighting its unique adaptations and their relevance to our work.

The Graphormer Encoder introduces graph-specific enhancements to address challenges inherent in graph-based data. Traditional Transformer Encoders excel at modeling global dependencies through self-attention mechanisms. However, they lack the ability to explicitly capture the spatial and relational structures unique to graph data. The Graphormer Encoder overcomes this limitation by integrating graph convolutional layers and graph-specific positional encoding mechanisms.

\begin{figure}[t]
\begin{center}
   \includegraphics[width=\columnwidth]{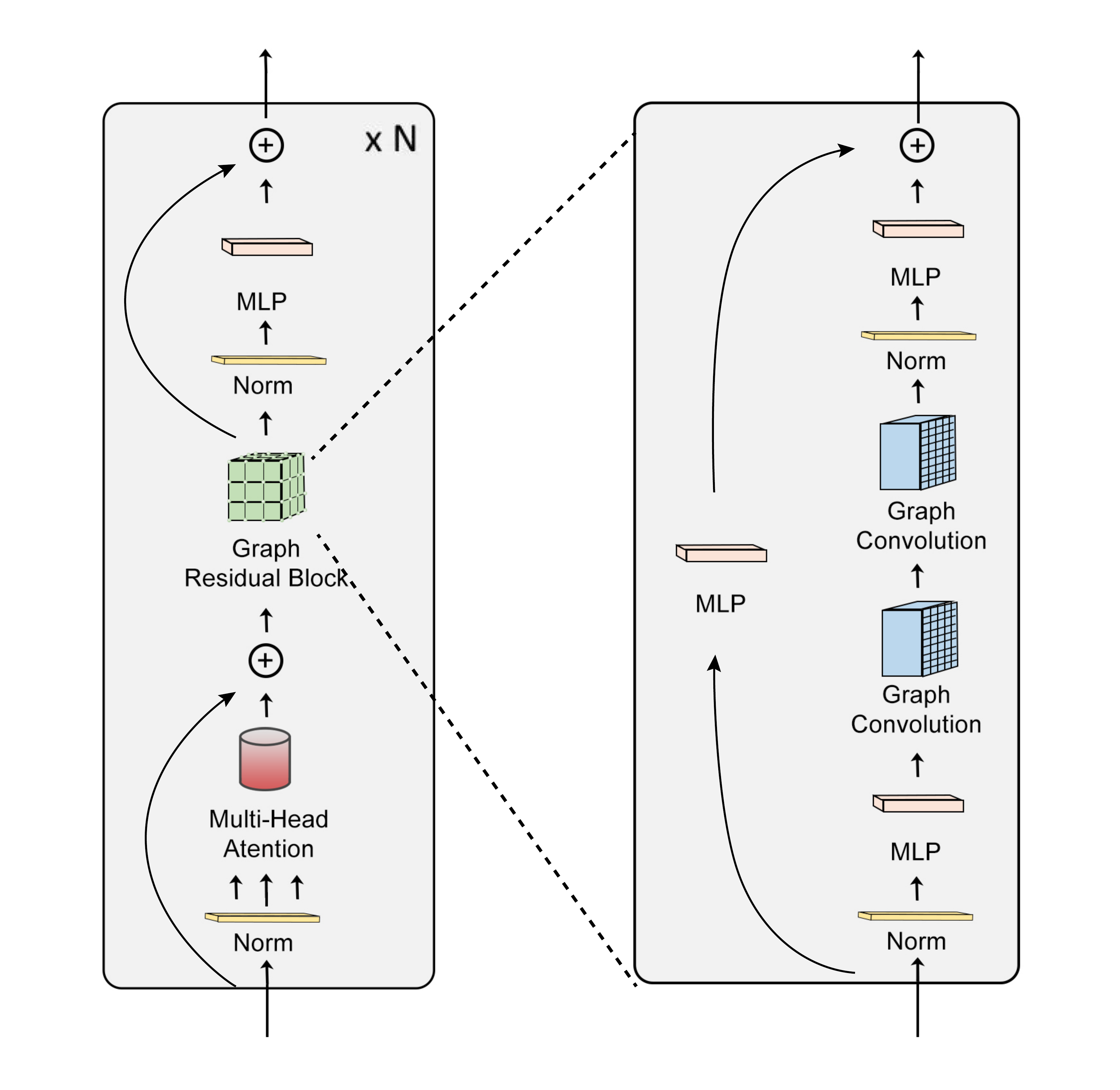}
\end{center}
   \caption{Architecture of a Graphormer Encoder. We introduce a transformer encoder augmented with graph convolutions to effectively capture both global and local interactions for 3D human mesh reconstruction. The encoder is composed of a series of four identical blocks.}
\label{fig:overview}
\end{figure}

As shown in Fig.~\ref{fig:overview}, the Graphormer Encoder consists of four identical blocks. Each block contains the following sequence of components: Layer Norm, Multi-Head Self-Attention (MHSA), Graph Residual Block, a second Layer Norm, and a Multi-Layer Perceptron (MLP). These components work together to model both global and local dependencies effectively. 

The MHSA module processes input sequences by transforming them into queries, keys, and values. These are then passed through parallel self-attention layers. The outputs from each layer are concatenated to form a contextual representation. This mechanism enables the Graphormer Encoder to capture global dependencies, similar to traditional Transformer Encoders. 

Additionally, the Graph Residual Block incorporates graph convolutional layers to enhance the modeling of fine-grained local interactions. These layers leverage spatial relationships between adjacent nodes or vertices in the graph. For tasks like 3D human mesh reconstruction and 3D contact prediction, this feature ensures accurate modeling of spatial correlations among mesh vertices. By combining MHSA and graph convolution, the Graphormer Encoder achieves superior performance in handling graph-based data, making it a highly effective foundation for our work.

\section{Detailed Experimental Description}


\subsection{Datasets}

\noindent\textbf{Datasets Used for Contact Point Prediction.}  
In this study, we employ three key datasets—\textbf{BEHAVE}, \textbf{RICH}, and \textbf{DAMON}—to train and evaluate our proposed method for contact point prediction and 3D human mesh reconstruction. Each dataset contributes uniquely to model development and evaluation, as detailed below.

The \textbf{BEHAVE} dataset, introduced by Bhatnagar et al., provides a comprehensive benchmark for human-object interactions in diverse and dynamic settings. It contains approximately 15,000 multi-view RGBD frames, covering various interaction scenarios relevant to gaming, virtual reality, and simulation. BEHAVE includes detailed annotations such as 3D SMPL body models, object fits, and precise human-object contact points. These features make it well-suited for evaluating both the accuracy of contact point predictions and the quality of 3D reconstructions. Its multi-view data further enables the robust assessment of methods that require spatial and interaction modeling.

The \textbf{RICH} dataset focuses on human-object interactions in domestic and household environments. Unlike BEHAVE, which emphasizes diverse scenarios, RICH highlights real-world contexts with detailed annotations of everyday activities and common object interactions. This dataset is particularly valuable for testing a model's generalizability to assistive robotics and smart home applications. Its close-range, frequent interactions within household contexts provide a complementary perspective, enhancing the robustness of our evaluation.

The \textbf{DAMON} dataset plays a dual role in this work. It is the primary resource for model training. DAMON provides dense annotations of 3D human-scene contact points, which are critical for learning pose-aware features. Its structured train-test split and diverse scenarios support robust model optimization and consistent performance evaluation.

\noindent\textbf{Datasets Used for 3D Human Reconstruction.} In addition to the datasets used for contact point prediction, we also leverage two widely used datasets—\textbf{3DPW} and \textbf{Human3.6M}—to evaluate the performance of our method on 3D human reconstruction metrics. Both datasets provide comprehensive benchmarks, ensuring reliable evaluation and fair comparison with SOTA methods.

The \textbf{3DPW} dataset focuses on real-world human poses captured in natural scenes. It contains approximately 60 video sequences featuring diverse activities, such as shopping, drinking coffee, and exercising, with precise annotations for 3D SMPL body models, 2D poses, and camera parameters. The dataset's emphasis on outdoor, in-the-wild scenarios makes it highly suitable for assessing the robustness of 3D human reconstruction algorithms in complex, unconstrained environments.

The \textbf{Human3.6M} dataset is a large-scale benchmark for 3D human pose estimation, featuring over 3.6 million frames of human actions captured in a controlled laboratory setting. It includes annotations for 3D joint positions, high-resolution videos, and calibrated camera views. This dataset is ideal for evaluating models under controlled conditions, providing a complementary perspective to the in-the-wild scenarios of 3DPW.

\subsection{Dataset Splits and Evaluation Protocols}  

In our GraphiContact contact perception experiments, we follow the same protocol as prior work by adopting the official (or source) train/test splits of each contact dataset and using a domain-specific training mixture depending on the evaluation target. Specifically, GraphiContact is trained using the combined training data from DAMON, RICH, and PROX, and is evaluated on DAMON-test and RICH-test. For the RICH-test setting, to isolate the effect of DAMON and ensure a fair, apples-to-apples comparison with BSTRO-style protocols, we train GraphiContact only on the RICH training split and report results on RICH-test. For the DAMON-test setting, we re-train GraphiContact using all available training data from RICH + PROX + DAMON, and evaluate on the DAMON test split; notably, DAMON inherits its train/test/val splits from the source HOT dataset rather than using a newly defined split. Finally, BEHAVE is used exclusively for out-of-domain generalization (i.e., test-only): we report performance on BEHAVE-test, and, following the dataset’s annotation constraints, we mask out foot–ground contacts during evaluation since BEHAVE does not provide ground-truth annotations for ground contact. 

In our GraphiContact reconstruction experiments, we follow the same widely adopted train/test protocols as prior work to ensure a fair and directly comparable evaluation. On Human3.6M, we use the standard subject-based split under Protocol \#2 (P2), training on subjects S1, S5, S6, S7, S8 and testing on S9, S11. Since ground-truth 3D mesh supervision for training is not publicly available for Human3.6M due to licensing constraints, we rely on the pseudo 3D mesh training supervision provided by prior work for mesh-level learning. On 3DPW, we train GraphiContact on the dataset's official training split and report results on the corresponding official test split, consistent with common practice in the literature to maintain comparability with established baselines.

\subsection{Implementation Details}  

The \textit{GraphiContact} model was implemented using the PyTorch framework and trained on an NVIDIA A100 GPU. PyTorch's dynamic computational graph and extensive library of functions supported rapid development and iterative optimization. The SIMU Strategy is incorporated to the robustness and uncertainty-awareness of 3D human contact point prediction and reconstruction from monocular images. To optimize model performance, we used the Adam optimizer with an initial learning rate of $8 \times 10^{-3}$ and a batch size of 4. A step-wise learning rate decay was employed to prevent overfitting, along with early stopping based on validation loss to ensure computational efficiency and robust convergence.

For efficiency, several lightweight modules in the architecture were simplified. Specifically, the image feature projection module and the fusion projection layer were each implemented with a single layer, and the sigmoid activation in the fusion projection layer was removed to streamline forward computation. Moreover, the Transformer encoder depth was set to 2 to balance representational capacity and computational overhead. During optimization, training was concentrated on the lightweight projection and fusion modules, allowing computational resources to be focused on the most influential components of the network.

Transfer Learning was adopted for the two transformer encoders to leverage complementary strengths from different datasets. The encoders were initialized with pretrained weights from the 3DPW and Human3.6M datasets. The outputs from these encoders were aggregated, with fixed contribution weights of 1 for 3DPW and 0.1 for Human3.6M, based on their domain relevance and data distribution characteristics. This weighting reflects the datasets' characteristics: 3DPW provides diverse, in-the-wild pose annotations, making it highly representative of real-world scenarios, while Human3.6M offers structured and precise 3D annotations in controlled environments. By emphasizing 3DPW, the model prioritizes robustness to real-world variability, while Human3.6M ensures accurate pose details.

Model predictions underwent a post-processing step where a confidence thresholding strategy was applied to binarize high-confidence outputs, thereby enhancing the interpretability of contact regions and improving the effectiveness of downstream tasks. Empirically, we observed that smaller contact areas, such as hands and elbows, tend to receive lower confidence scores, indicating that threshold selection can affect the inclusion of such regions. Although the threshold was not quantitatively optimized, this heuristic demonstrated practical utility.

Finally, we evaluate the practical efficiency of GraphiContact on the DAMON benchmark~\cite{tripathi2023deco} by reporting both single-image inference latency and backbone parameter count. As summarized in Table~\ref{table:Inference Time}, GraphiContact operates at an $\mathcal{O}(10^2)$ ms-per-image latency and an $\mathcal{O}(10^2)$M parameter scale, indicating that our approach is not an excessively large model and remains far from second-level runtimes. This efficiency supports practical single-branch deployment in interactive settings, while preserving the strong contact prediction performance reported in the main paper.

\begin{table}[t]
\centering
\caption{Backbone complexity and single-image inference latency for contact prediction on DAMON.}
\vspace{-0.5em}
\label{table:Inference Time}

\scriptsize 
\setlength{\tabcolsep}{5pt} 
\renewcommand{\arraystretch}{1.15} 

\begin{tabular}{c|c|c}  
\Xhline{1.2pt}
\bf Methods & \bf Params (M) $\downarrow$ & \bf Inference time (ms/img) $\downarrow$ \\
\hline
\rowcolor{gray!13}BSTRO~\cite{huang2022capturing} & \underline{$25$}  & \underline{$9$}  \\
POSA~\cite{hassan2021populating}                  & $\textbf{2}$ & $\textbf{2}$ \\
\rowcolor{gray!13}DECO~\cite{tripathi2023deco}    & $87$  & $20$  \\
LEMON~\cite{yang2024lemon}                        & $29$  & $18$  \\
\rowcolor{gray!13}CONTHO~\cite{nam2024joint}      & $141$ & $25$  \\
PICO~\cite{PICO}                                  & $-$   & $2498$ \\
\rowcolor{gray!13}GraphiContact (Ours)            & $128$ & $332$ \\
\Xhline{1.2pt}
\end{tabular}

\vspace{2mm}
\textbf{Bold} indicates the best performance for each metric, while \underline{underline} denotes the second-best result.
\vspace{1mm}
\end{table}

\subsection{Qualitative Comparison}  

\begin{figure*}[!t]
   \centering
   \includegraphics[width=0.8\textwidth]{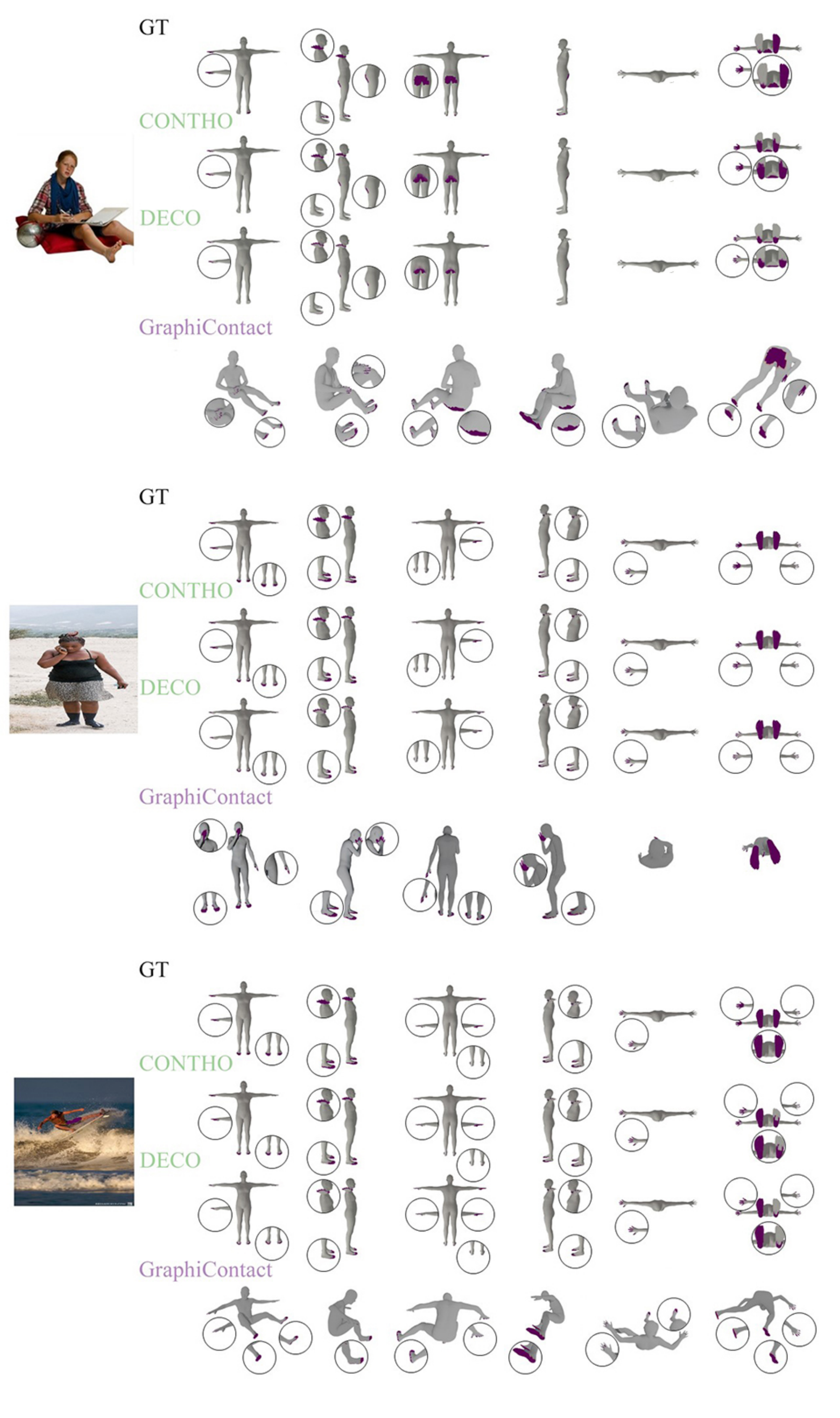}
   \vspace{0pt}
   \caption{Qualitative Comparison of Contact Point Prediction and 3D Mesh Reconstruction (Scenario A)}
\vspace{-15pt}
\label{fig:results visualization1}
\end{figure*}

\begin{figure*}[!t]
   \centering
   \includegraphics[width=0.8\textwidth]{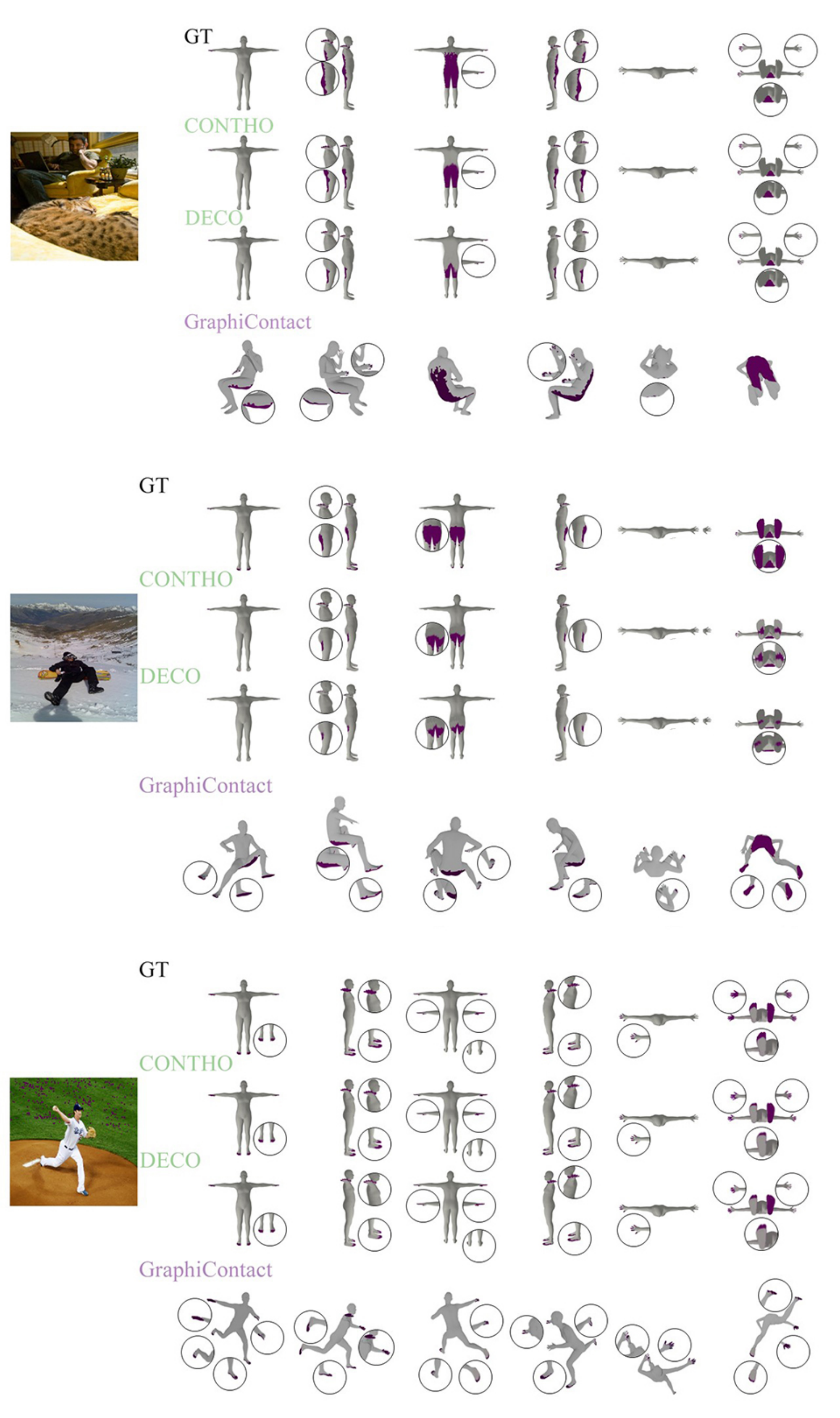}
   \vspace{-10pt}
   \caption{Qualitative Comparison of Contact Point Prediction and 3D Mesh Reconstruction (Scenario B)}
\vspace{-15pt}
\label{fig:results visualization2}
\end{figure*}

\begin{figure*}[!t]
   \centering
   \includegraphics[width=0.8\textwidth]{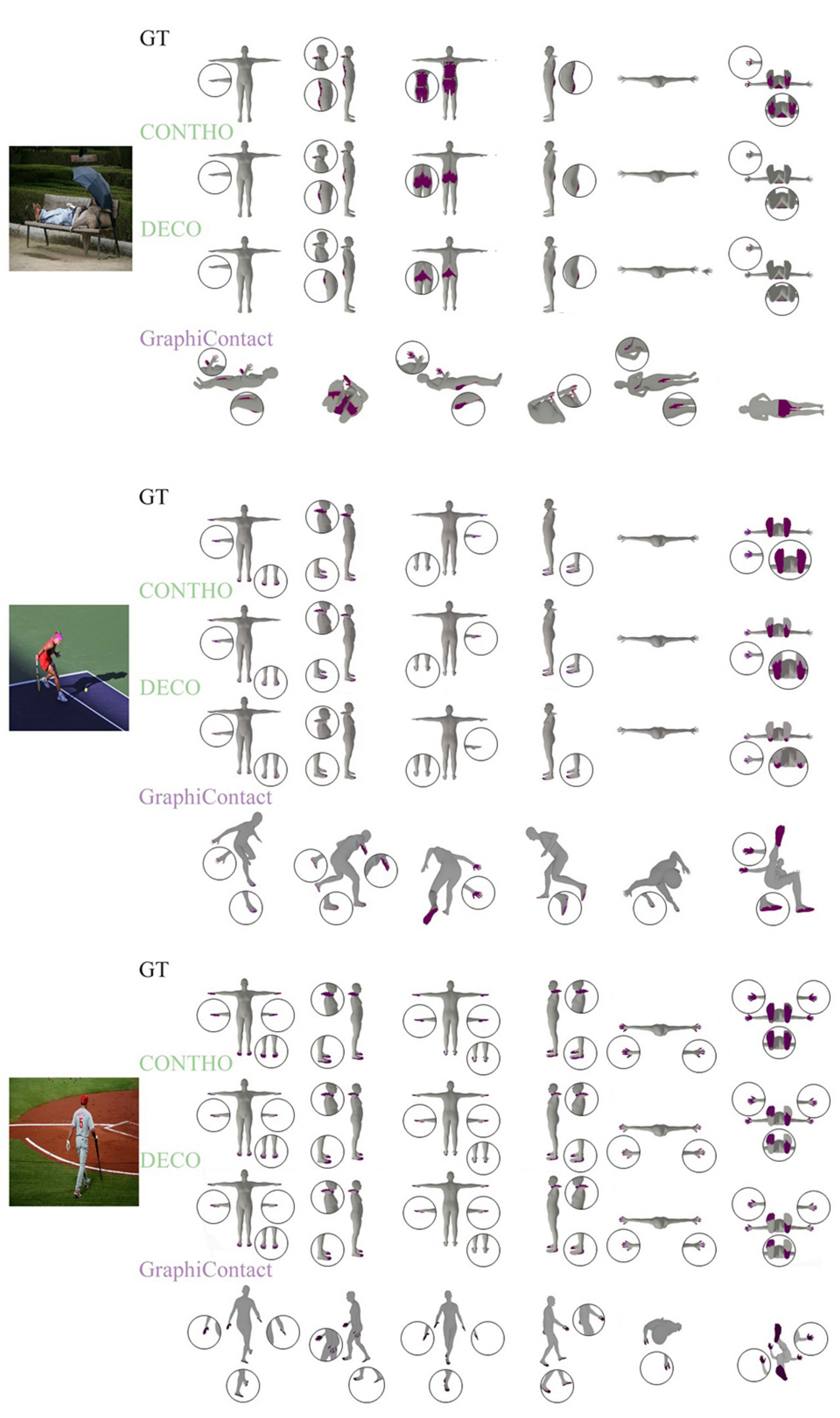}
   \vspace{-10pt}
   \caption{Qualitative Comparison of Contact Point Prediction and 3D Mesh Reconstruction (Scenario C)}
\vspace{-15pt}
\label{fig:results visualization3}
\end{figure*}

\begin{figure*}[!t]
   \centering
   \includegraphics[width=0.8\textwidth]{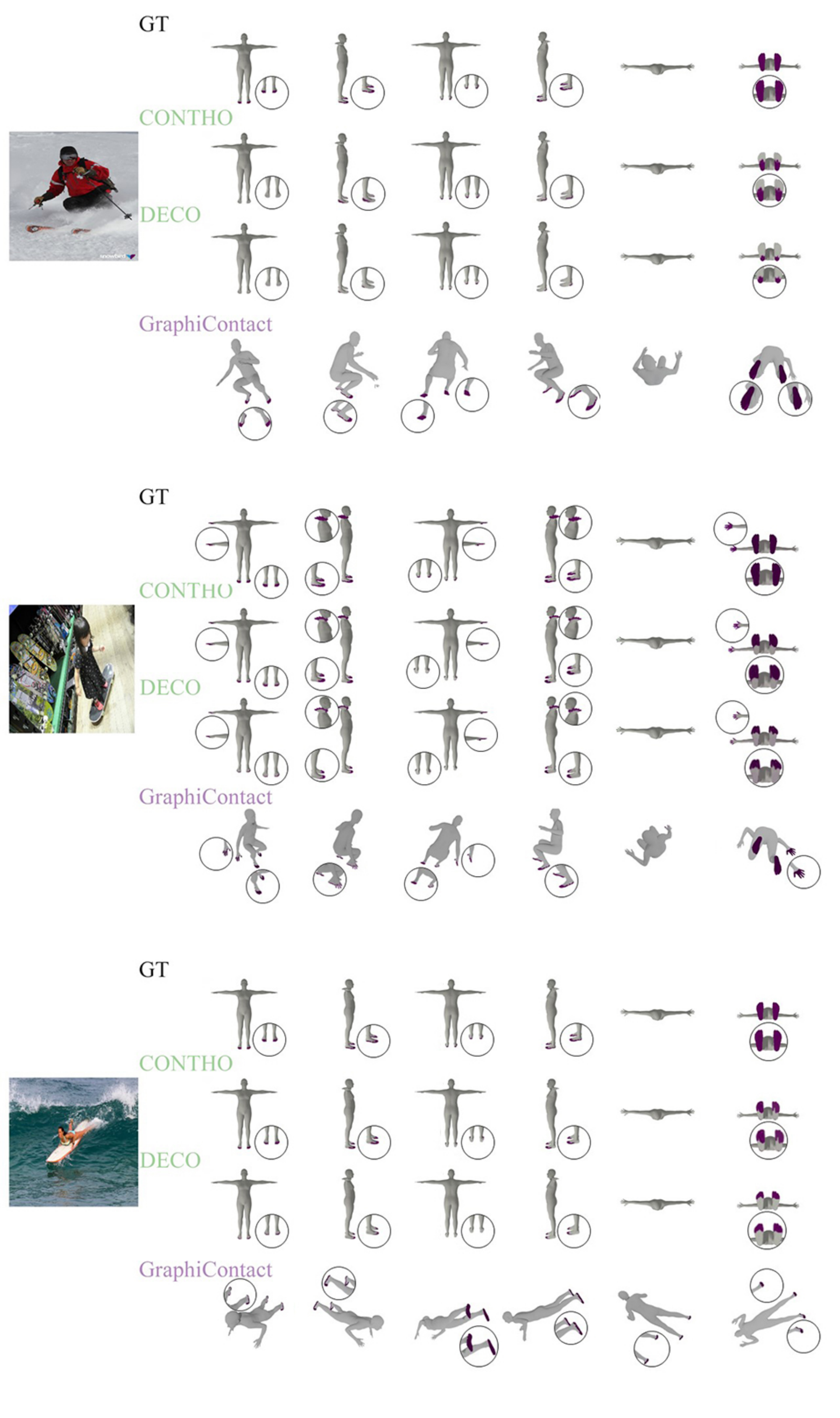}
   \vspace{-10pt}
   \caption{Qualitative Comparison of Contact Point Prediction and 3D Mesh Reconstruction (Scenario D)}
\vspace{-15pt}
\label{fig:results visualization4}
\end{figure*}

Fig.~\ref{fig:results visualization1}, Fig.~\ref{fig:results visualization2}, Fig.~\ref{fig:results visualization3}, and Fig.~\ref{fig:results visualization4} provide additional qualitative comparisons with the DECO~\cite{tripathi2023deco} and CONTHO~\cite{nam2024joint} method, supplementing the results discussed in the main paper. These examples clearly demonstrate the dual advantages of our method in capturing detailed human-environment interactions and reconstructing realistic human poses. The additional visualizations included in these figures further highlight the robustness of our model across diverse scenarios, offering a more comprehensive understanding of its performance advantages.

\section{Limitations and Future Work}

While GraphiContact demonstrates strong performance in 3D human mesh reconstruction and contact point prediction, there are a few limitations that need to be addressed for broader applicability.

One challenge is that our current approach primarily handles single-image uncertainty, lacking temporal context. This limits its ability to model dynamic human-environment interactions, such as motion continuity or velocity-aware contact forces. Future work could integrate temporal models like video-based transformers to improve dynamic interaction understanding.

Additionally, the fixed topology of the SMPL model limits adaptability to extreme poses or non-rigid scenes. Enhancing the flexibility of the model and improving generalization across diverse environments, such as cluttered backgrounds, will be key for real-world applications, including AR/VR and robotics.

Moreover, for HCI applications, our model would benefit from further refinement in real-time performance, particularly in fast-changing or interactive environments, such as assistive robotics or immersive AR experiences. Addressing these challenges will enhance the system’s ability to seamlessly integrate into interactive systems, enabling more responsive and adaptive human-computer interactions.

\section{Broader Context on Trustworthy and Controllable Visual Intelligence}

Beyond the immediate scope of human--scene contact prediction, recent visual intelligence research increasingly focuses on controllability, robustness, and trustworthiness under manipulated inputs. In generative modeling, prior work studies concept erasure, recovery, and denoising-time control across diffusion and flow-based models~\cite{lu2024mace,gao2024eraseanything,gao2025revoking,li2025set}, while robust watermarking benchmarks expose vulnerability to modern editing pipelines~\cite{lu2024robust}. Related efforts on image manipulation further examine training-free composition, physically plausible FLUX-based composition, and region-based drag editing~\cite{lu2023tf,lu2025does,zhou2025dragflow}.

Related concerns also span visual security, content authenticity, multimodal understanding, and reliable inference. Prior work addresses copy-move forgery detection, secure secret embedding in 3D Gaussian Splatting, and speech deepfake detection~\cite{lu2022copy,ren2025all,xuan2025wavesp}. In multimodal reasoning, A-MESS, S$^2$-KD, and recent LVLM hallucination mitigation improve semantic synchronization, knowledge transfer, and saliency-aware reliability control~\cite{shen2025mess,wang2025s,zhang2026hallucination}. Trustworthy inference is further studied in graph and temporal settings, including fraud detection, anomaly detection, and event-based human understanding~\cite{zhang2025dual,zhang2025dconad,zhang2025frect,yang2025temporal}.

Structured perception and system-level trustworthiness form another important axis, spanning lane topology understanding, autonomous-driving detection, fairness and governance in AI, and numerical stability in iterative systems~\cite{li2025reusing,lin2025butter,lin2024comprehensive,jiang2025never,lin5074292insertion}. Although these directions do not directly target monocular vertex-level human--scene contact prediction, they collectively highlight a broader pursuit of reliable, controllable, and trustworthy intelligence under real-world constraints. Our work connects to this broader context from the discriminative 3D understanding side, by focusing on robust contact inference under occlusion and noise.

\end{document}